\documentclass[12pt]{article}

\usepackage{amsthm}
\usepackage{amsmath}
\usepackage{lineno,hyperref}
\usepackage{epsfig}
\usepackage{graphicx}
\usepackage{subfigure}
\usepackage{natbib}
\usepackage{multirow}
\usepackage{booktabs}
\usepackage{threeparttable}
\usepackage{color}
\usepackage{enumerate}
\usepackage{url} 

\usepackage{amsmath}
\newtheorem{theorem}{Theorem}
\newtheorem{definition}{Definition}
\newtheorem{lemma}{Lemma}
\newtheorem{assumption}{Assumption}

\usepackage{amssymb}
\usepackage{appendix}
\usepackage{authblk}

\usepackage{algorithmic}
\usepackage[ruled,vlined]{algorithm2e}

\newcommand{\ve}[1]{\mbox{\boldmath ${#1}$}}
\newcommand{\vesub}[2]{\mbox{{\boldmath ${#1}$}$_{#2}$}}
\newcommand{\vesup}[2]{\mbox{{\boldmath ${#1}$}$^{#2}$}}
\newcommand{\vess}[3]{\mbox{{\boldmath ${#1}$}$_{#2}^{#3}$}}

\newcommand{\var}{{\rm var}}

\newcommand{\CC}[2]{\tbinom{#1}{#2}}

\begin{document}
\thispagestyle{empty} 

\def\spacingset#1{\renewcommand{\baselinestretch}%
{#1}\small\normalsize} \spacingset{1}




\title{\bf DeepSuM: Deep Sufficient Modality Learning Framework}
  \author{Zhe Gao \hspace{.2cm}\\
    School of Management, University of Science and Technology of China\\
    and\\
    Jian Huang  \\
    Department of Data Science and Artificial Intelligence, The Hong Kong Polytechnic University \\
    and\\
    Ting Li  \\
    Department of Applied Mathematics, The Hong Kong Polytechnic University \\
    and\\
    Xueqin Wang  \\
    School of Management, University of Science and Technology of China \\
    }
  \maketitle


\begin{abstract}
Multimodal learning has become a pivotal approach in developing robust learning models with applications spanning multimedia, robotics, large language models, and healthcare. The efficiency of multimodal systems is a critical concern, given the varying costs and resource demands of different modalities. This underscores the necessity for effective modality selection to balance performance gains against resource expenditures. In this study, we propose a novel framework for modality selection that independently learns the representation of each modality. This approach allows for the assessment of each modality's significance within its unique representation space, enabling the development of tailored encoders and facilitating the joint analysis of modalities with distinct characteristics. Our framework aims to enhance the efficiency and effectiveness of multimodal learning by optimizing modality integration and selection.



\end{abstract}


\spacingset{1.9} 


\section{Introduction}\label{intro}

Multimodal learning involves integrating and processing data from multiple sensory channels or modalities, such as text, images, and audio, to build more robust and comprehensive learning models. 
It has widespread applications in multimedia \citep{Naphade2006,atrey2010multimodal,dang2017multimodal}, robotics \citep{kirchner2019embedded,lee2019making}, large language model \citep{huang2023language,gao2023llama,driess2023palm} and healthcare \citep{muhammad2021comprehensive,vanguri2022multimodal,lipkova2022artificial}.

While extensive studies have been conducted on the applications of multimodal learning, theoretical investigations remain relatively limited. \cite{sun2020} introduced an information-theoretic framework to elucidate semi-supervised multimodal learning. \cite{huang2021makes} further developed a broader concept based on latent space representation, offering insights into why multimodal learning may surpass unimodal learning in effectiveness.

However, multimodal learning sometime fails. In \cite{du2023unimodalfeaturelearningsupervised,wang2020makes,mbintvqa}, authors pointed out the problem of modality laziness, a phenomenon where some modalities appear more dominant than others during multimodal learning. \cite{wang2020makes} found that different modalities have different convergence rates, making the jointly trained multimodal model fail to match or outperform its unimodal counterpart. 
To tackle the problem, \cite{huang2022modality} introduced modal competition characterizes under a specific data distribution. 

Another important consideration in multimodal learning is efficiency. In practice, one must account for the cost of different modalities, which can vary significantly in terms of data acquisition, processing requirements, and storage needs. For example, collecting and processing high-resolution video data is often more resource-intensive than handling text or audio data. Therefore, it's crucial to evaluate the trade-offs between the benefits of incorporating additional modalities and their associated costs. This involves assessing whether the performance improvements gained from additional modalities justify the increased resource expenditure. 

Moreover, some modalities can negatively impact the overall system. For example, multimodal medical data often includes uninformative features due to suboptimal data collection, such as noise in multiomics data \citep{argelaguet_2020}, varying quality of histopathological images \citep{yagi2011color}, and complex missing patterns in tabular data \citep{yelipe2018efficient}. This highlights the importance of assessing the informativeness of each feature and modality. Often, fewer modalities may suffice to achieve the desired performance, and the marginal benefit of adding new modalities decreases as more are included. To improve the efficiency of multimodal learning, it's crucial to evaluate and select the most useful modalities while discarding less useful ones. All above motivate us to develop a comprehensive method for modality selection and learning.

The literature on modality selection is relatively limited. \cite{han2022multimodal} introduce the concept of True-Class-Probability to quantify the classification confidence across different modalities and propose a dynamic fusion strategy for classification problems. \cite{he2024} develop the Marginal Contribution Feature Importance, based on the Shapley value, and present efficient modality selection algorithms using submodular optimization. However, these methods generally assume that the spaces of different modalities are either identical or similar, and they do not address the association between different modalities at the feature representation level.

In this study, we propose a novel framework that integrates modality selection. Our approach begins by learning the representation of each modality independently. This allows us to evaluate the significance of each modality within its representation, facilitating the design of tailored encoders or representation learning models for each modality. This methodology enables the joint analysis of modalities that occupy distinctly different spaces.

More specifically, adopting the idea of sufficient representation,
we learn a non-parametric representation of each modality using a neural network. The importance of each modality is assessed based on the dependency measure between its representation and the response variable. We ensure independence among representations across different modalities, preventing redundant information from one modality from being integrated into learning process. This approach enhances the efficiency of the learning process and significantly improves the model's interpretability.

Our model not only supports learning multiple representations but also incorporates various modality selection strategies. Initially, for a model with no preselected modalities, we assess the utility of each candidate modality for downstream tasks by comparing their dependency measure. Subsequently, after some modalities have been preselected, we evaluate the potential addition of a new modality by checking if its dependency metric equals $0$. This approach also allows us to filter out non-contributive modalities when multiple candidates are considered simultaneously.

Our contributions are threefold. Firstly, we introduce a multi-modal learning method that leverages feature representation and sufficient dimension reduction, adaptable across diverse modality spaces. Secondly, 
we provide a joint low-dimensional representation of multimodal data. Finally, we develop modality selection techniques to enhance model performance. Moreover, we provide a detailed analysis of the algorithm's convergence and the theoretical guarantee of our modality selection methods, demonstrating the superiority of our approach in specific scenarios.

The remainder of this paper is organized as follows: Section 2 introduces the multi-modal learning framework and algorithm and modality selection procedure. Section 3 delves into the theoretical results of the method. Section 4, 5 showcases case studies to validate and demonstrate the practical utility of the proposed method. Section 6 offers concluding remarks, encapsulating our main contributions.
\section{Methodology}

\subsection{Notation}
Assume the data is given by $\ve{X} = (X^{(1)}, X^{(2)},  \dots, X^{(K)})$, which consists of $K$ modalities. The domain of the $k$-th modality is $X^{(k)} \in \mathcal{X}^{(k)} \subseteq \mathbb{R}^{p_k}$. Denote $\mathcal{X} = \mathcal{X}^{(1)} \times \cdots \times \mathcal{X}^{(K)}$. Let $Y \in \mathcal{Y}$ be the target variable. Assume $\ve{Z} \in \mathcal{Z}$ is the latent variable, with $g^{*}: \mathcal{X} \rightarrow \mathcal{Z}$ as the true mapping from the input space. Additionally, we define $h^{*}: \mathcal{Z} \rightarrow \mathcal{Y}$ as the true mapping for target variable.
Consider an i.i.d. samples $\{(\vesub{X}{i}, Y_i), i = 1, 2, \dots, n\}$ from an unknown distribution $\mathcal{D}$ such that
$$\mathbb{P}_{\mathcal{D}}(\ve{X}, Y) = \mathbb{P}_{Y \mid \mathbf{X}}\left(Y \mid h^{*} \circ g^{*}(\ve{X})\right) \mathbb{P}_{\mathbf{X}}(\ve{X})$$
Here $h^{*} \circ g^{*}(\ve{X})=h^{*}\left(g^{*}(\ve{X})\right)$ represents the composite function of $h^{*}$ and $g^{*}$.

Given the data set, we define the loss function of the target variable as $\ell(\cdot, \cdot)$. Our goal is to find $h$ and $g$ to minimize the empirical risk:
\begin{align*}
\min & \quad \hat{\mathcal{L}}\left(h \circ g\right) \triangleq \frac{1}{n} \sum_{i=1}^n \ell\left(h \circ g\left(\vesub{X}{i}\right), Y_i\right) \\
\text { subject to } & \quad h \in \mathcal{H}, g \in \mathcal{G} .
\end{align*}
Here $\mathcal{H}, \mathcal{G}$ are the function classes. The population risk is defined as
\begin{align*}
    \mathcal{L}\left(h \circ g\right) \triangleq \mathbb{E}_{(\mathbf{X}, Y) \sim \mathcal{D}}\big[\ell\left(h \circ g\left(\ve{X}\right), Y\right)\big]. 
\end{align*}

\subsection{The Deep Sufficient Modality Learning Framework
}
In this subsection, we introduce our Deep Sufficient Modality Learning Framework (DeepSuM), focusing initially on establishing the latent space mapping, denoted as $g$. Given that the dataset $\ve{X}$ encompasses $K$ modalities, it typically presents a complex structure. Our objective is to develop a representation of $\mathcal{X}$ that embodies three key attributes: sufficiency, low dimensionality, and disentanglement. 
We start with the concept of sufficient representation firstly proposed in \citep{cook1998regression}, and we consider its nonparametric generalization.

\begin{definition}\label{ddr}
    A measurable function $g: \mathcal{X} \rightarrow \mathbb{R}^d$ is said to be a sufficient representation of $X$ if
$$
Y \perp X \mid g(X),
$$
that is, $Y$ and $X$ are conditionally independent given $g(X)$. 
\end{definition}
The condition holds if and only if the conditional distribution of $Y$ given $X$ are equal to the conditional distribution of $Y$ given $g(X)$. Consequently, the information that $X$ provides about $Y$ is fully captured by $g(X)$.

Denote the class of sufficient representations satisfying by
$$
 \mathcal{G}_0=
\left\{g: \mathcal{X} \rightarrow \mathbb{R}^d, g \text { satisfies } Y \perp \ve{X} \text { given } g(\ve{X})\right\}.
$$
Here we assume the latent space is a $d$-dimensional Euclidean space. 

Since $\ve{X}$ contains $k$ modalities, we use the early fusion strategy to contain the different modalities. We assign a mapping to each modality, then the latent mapping can be expressed as $g = g_1 \oplus g_2 \oplus \cdots \oplus g_K$. We need to find the sufficient representations for each modality. Thus the mapping set of the $k$-th modality becomes
$$
\mathcal{G}_k =\left\{g_k: \mathcal{X}^{(k)} \rightarrow \mathbb{R}^{d_k}, g_k \text { satisfies } Y \perp \ve{X}^{(k)} \text { given } g_k(\ve{X}^{(k)})\right\}.
$$
For an injective measurable transformation $T : \mathbb{R}^{d_k} \rightarrow \mathbb{R}^{d_k}$ and a function $g_k \in \mathcal{G}_k$, the composition $T \circ g_k$ also belongs to $\mathcal{G}_k$. This result comes from the fact that an injective measurable mapping preserves conditional independence, thereby ensuring that the class $\mathcal{G}_k$ is invariant under such transformations.
Based on this invariance, we further add the assumption of distribution to sufficient representation and simplify its form.
Considering a invertible matrix $R$, the transformation $R g_k$ remains within $\mathcal{G}_k$. This property allows us to rescale $g_k$ such that it possesses an identity covariance matrix. Drawing upon the Maxwell characterization of Gaussian distributions \citep{Maxwell1860}, we can reformulate the class $\mathcal{G}_k$ to include these normalized transformations,
$$
\mathcal{G}_k =\left\{g_k: \mathcal{X}^{(k)} \rightarrow \mathbb{R}^{d_k},   g_k(\ve{X}^{(k)}) \sim N(0, I_{d_k})\right\}.
$$
The existence of such a representation under mild conditions is shown in \citep{huang2024deep}. 

After integrating different modalities through early fusion techniques, we additionally introduce independence assumptions to ensure that each modality more effectively encapsulates its information.  Therefore, our sufficient representation needs to meet the following conditions:
\begin{equation}
\ve{X} \perp Y \mid g(\ve{X}) \text { and } g(\ve{X}) \sim N\left(\mathbf{0}, I_d\right), g_k(\vesup{X}{(k)}) \perp g_l(\vesup{X}{(l)}), \forall 1 \leq k < l \leq K. 
\label{opt0}
\end{equation}
The function class of sufficient representations is
$$
\mathcal{G} =\left\{g: g_1 \oplus g_2 \oplus \cdots \oplus g_K: g_k \in \mathcal{G}_k,  g_k(\vesup{X}{(k)}) \perp g_l(\vesup{X}{(l)}), \forall 1 \leq k < l \leq K \right\}.
$$

Next, we construct the nonparametric estimation of the sufficient representation. For the $k$-th modality, we introduce a dependence measure $\mathcal{V}$ to describe the condition dependence in Definition~\ref{ddr}. Specifically, we assume $\mathcal{V}$ satisfies the following properties: 

(a) $\mathcal{V}[X, Y] \geq 0$ with $\mathcal{V}[X, Y]=0$ if and only if $X \perp Y$; 

(b) $\mathcal{V}[X, Y] \geq$ $\mathcal{V}[g(X), Y]$ for all measurable function $g$; 

(c) $\mathcal{V}[X, Y]=$ $\mathcal{V}\left[g^*(X), Y\right]$ if and only if $g^* \in \mathcal{G}$. 


These properties imply that we can obtain the mapping of the $k$-th modality by minimizing $\mathcal{V}[g_k(\vesup{X}{(k)}), Y]$. For the normality of $g_k(\vesup{X}{(k)})$, we use divergence measure $\mathbb{D}$ which satisfies
$\mathbb{D}(\mu(g_k(\vesup{X}{(k)})), \gamma_{d_k}) \geq 0$ for every measurable function $g_k$ and $\mathbb{D}(\mu(g_k(\vesup{X}{(k)})), \gamma_{d_k}) = 0$  if and only if $g_k(\vesup{X}{(k)})$ follows $N(0, I_{d_k})$. Here we use $\gamma_d$ denotes measure induced by $N(0, I_{d_k})$, $\mu(g_k)$ is the measure induced by $g_k$. Thus, the latent representation of $k$-th modality can be obtained by the following constrained minimization problem:
\begin{align}\label{optk}
 \operatorname{argmin}_{g_k \in \mathcal{G}_k} &-\mathcal{V}[g_k(\vesup{X}{(k)}), Y],  \\
 \text { subject to } & \mathbb{D}(\mu(g_k(\vesup{X}{(k)})), \gamma_{d_k}) = 0 \nonumber .
\end{align}
The Lagrangian form of this minimization problem is
$$
F_k(g_k) = -\mathcal{V}[g_k(\vesup{X}{(k)}), Y]+\lambda_k \mathbb{D}(\mu(g_k(\vesup{X}{(k)})), \gamma_{d_k}),
$$
where $\lambda_k \geq 0$ is a tuning parameter. By combining the results of $K$ modalities and the independence assumption, we obtain our sufficient representation optimization problem:
\begin{align}\label{optall}
 \operatorname{argmin}_{g} &- \sum_{k=1}^K \mathcal{V}[g_k(\vesup{X}{(k)}), Y],   \\
 \text { subject to } & \mathbb{D}(\mu(g_k(\vesup{X}{(k)})), \gamma_{d_k}) = 0, k = 1, \dots, K, \nonumber \\
 & \mathcal{V}[g_k(\vesup{X}{(k)}), g_l(\vesup{X}{(l)})] = 0, \forall 1 \leq k < l \leq K. \nonumber
\end{align}
The Lagrangian form of this minimization problem is
$$
F(g) = -\sum_{k=1}^K \big[\mathcal{V}[g_k(\vesup{X}{(k)}), Y]  +\lambda_k \mathbb{D}(\mu(g_k(\vesup{X}{(k)})), \gamma_{d_k}) \big] + \sum_{1 \leq k < l \leq K} \xi_{kl}\mathcal{V}[g_k(\vesup{X}{(k)}), g_l(\vesup{X}{(l)})],
$$
with parameters $\xi_{kl}$, $\forall 1 \leq k < l \leq K$.

After constructing the representation of the latent variables, we can proceed to the construction of the downstream model. The predict function can be obtained by the following optimization problem:
\begin{align}
    \hat{h} = \operatorname{argmin}_{h \in \mathcal{H}} \mathcal{L}\left(h \circ g\right) \triangleq \mathbb{E}_{(\mathbf{X}, Y) \sim \mathcal{D}}\big[\ell\left(h \circ g\left(\mathbf{X}\right), Y\right)\big]. 
\end{align}


\subsection{The DeepSuM Algorithm}
We now show how to solve optimization problems through samples. 
For the independence measure, we use the 
distance covariance \citep{dc2007}, 
which is defined as
\begin{align*}
    \mathcal{V}[Z, Y]=&\frac{1}{c_p c_q} \int_{\mathbb{R}^{p+q}} \frac{\left|\varphi_{Z, Y}(s, t)-\varphi_Z(s) \varphi_Y(t)\right|^2}{|s|_p^{1+p}|t|_q^{1+q}} d t d s \\
    = & \mathbb{E}\left[\left\|Z-Z^{\prime}\right\|\left\|Y-Y^{\prime}\right\|\right]+\mathbb{E}\left[\left\|X-X^{\prime}\right\|\right] \mathbb{E}\left[\left\|Y-Y^{\prime}\right\|\right]] \\
& -2 \mathbb{E}\left[\left\|Z-Z^{\prime}\right\|\left\|Y-Y^{\prime \prime}\right\|\right]
\end{align*}
where $c_p, c_q$ are some constants and $\varphi$ is the characteristic function. The empirical form of the distance covariance is
\begin{align*}
    \mathcal{V}_n[Z, Y] = & \frac{1}{\CC{n}{2}} \sum_{1 \leq i < j \leq n}  \left\|Z_i-Z_j\right\|\left\|Y_i-Y_j\right\| + \frac{1}{\CC{n}{2}} \sum_{1 \leq i < j \leq n}  \left\|Z_i-Z_j\right\| \frac{1}{\CC{n}{2}} \sum_{1 \leq i < j \leq n}\left\|Y_i-Y_j\right\| \\
    & -  \frac{2}{\CC{n}{3}} \sum_{i, j, u}  \left\|Z_i-Z_j\right\|\left\|Y_i-Y_u\right\|.
\end{align*}

For the divergence measure $\mathbb{D}$, we use $f$-divergence \citep{ali1966general} for $\mu << \gamma$ defined as
\begin{align*}
    \mathbb{D}_f(\mu \| \gamma)=\int_{\mathbb{R}^d} f\left(\frac{\mathrm{d} \mu}{\mathrm{d} \gamma}\right) \mathrm{d} \gamma
\end{align*}
where $f: \mathbb{R}^{+} \rightarrow \mathbb{R}$ is a differentiable convex function satisfying $f(1) = 0$. The $f$-divergence admits the following variational formulation \citep{keziou2003dual}.
\begin{lemma}
    Suppose that $f$ is differentiable, proper, convex and lower-semicontinuous on its domain. Then,
$$
\mathbb{D}_f(\mu \| \gamma)=\max _{D: \mathbb{R}^d \rightarrow \operatorname{dom}\left(f^*\right)} \mathbb{E}_{Z \sim \mu} D(Z)-\mathbb{E}_{W \sim \gamma} f^*(D(W)),
$$
where $f^*$ is the Fenchel conjugate. In addition, the maximum is attained at $D(\mathbf{z})=f^{\prime}\left(\frac{\mathrm{d} \mu}{\mathrm{d} \gamma}(\mathbf{z})\right)$.
\end{lemma}

To obtain the estimate of $\mathbb{D}_f(\mu \| \gamma)$, we need to estimate an optimal discriminator $D_\phi$ approximating the optimal dual function $D(\mathbf{z})=f^{\prime}(\frac{\mathrm{d} \mu}{\mathrm{d} \gamma}(\mathbf{z}))$. Following the idea in \citep{huang2024deep}, we use the residual maps $\mathbb{T}(\mathbf{z})=\mathbf{z}+s \mathbf{v}(\mathbf{z})$ with a small step size $s>0$ that most decreases the $f$-divergence, to push the samples to the target distribution. Based on the deep density ratio estimation, we construct the estimator as
\begin{align*}
\widehat{D}_\phi \in \arg \min _{D_\phi} \frac{1}{n} \sum_{i=1}^n\left\{\log \left[1+\exp \left(D_\phi\left(Z_i\right)\right)\right]+\log \left[1+\exp \left(-D_\phi\left(W_i\right)\right)\right]\right\}
\end{align*}
where $W_i \sim \gamma_d, i=1,2, \ldots, n$ and $Z_i, i=1,2, \ldots, n$ is the sufficient representation.
The problem is solved by stochastic gradient descent (SGD). Then the estimated density ratio $\hat{r}(\mathbf{z})=\exp \left(-\widehat{D}_\phi(\mathbf{z})\right)$, which is well estimated by $\widehat{D}_\phi$ \citep{huang2024deep}. 
Thus, after determining the discriminator, the loss function of the sufficient representation for the $k$-th modality is
\begin{align*}
     -\mathcal{V}_n[g_k(\vesup{X}{(k)}), Y] + \lambda_k \frac{1}{n} \sum_{i=1}^n\left\|g_k(\vess{X}{i}{(k)})-W_i\right\|^2 .
\end{align*}



Integrating all modalities, the DeepSuM algorithm can be summarized as following:
\begin{algorithm}[h!]
\renewcommand{\algorithmicrequire}{\textbf{Input:}}
\renewcommand{\algorithmicensure}{\textbf{Output:}}
\footnotesize
\caption{\quad DeepSuM Algorithm}
\label{alg:cap}
\begin{algorithmic}[1]
    \REQUIRE  $\{(\vesub{X}{i}, Y_i)\}_{i = 1}^n$, tuning parameters $\{\lambda_i\}_{i=1}^K$, $\{\xi_{kl}\}_{1 \leq k < l \leq K}$, dimension of latent space $\{d_i\}_{i=1}^K$, step parameters $\{s_i\}_{i=1}^K$, sample $\{W_i^{(k)}\}_{i=1}^n \sim \gamma_{d_k}, k = 1, \dots, K$;
    \FOR{$k= 1$ \TO $K$} 
    \STATE Let $Z_{ik} = g_k(\vesup{X_i}{(k)}), i = 1, \dots, n$;
    
    Solve
    $$
    \widehat{D}_\phi^{k} \in \arg \min _{D_\phi} \frac{1}{n} \sum_{i=1}^n\left\{\log \left[1+\exp \left(D_\phi\left(Z_{ik}\right)\right)\right]+\log \left[1+\exp \left(-D_\phi\left(W_i^{(k)}\right)\right)\right]\right\} .
    $$ 
    
    Define the residual map $\mathbb{T}(\mathbf{z})=\mathbf{z}-s_k \nabla f^{\prime}(\hat{r}(\mathbf{z}))$ with $\hat{r}(\mathbf{z})=\exp \left(-\widehat{D}_\phi(\mathbf{z})\right)$.
    
    Update the particles $Z_{ik}=\mathbb{T}\left(Z_{ik}\right), i=1,2, \ldots, n$.
    \STATE Update $g_k$ via minimizing 
    \begin{align*}
        - \mathcal{V}_n[g_k(\vesup{X}{(k)}), Y] +  \frac{\lambda_k}{n}\sum_{i = 1}^n ||g_k(\vesup{X_i}{(k)}) - Z_{ik}||^2 +\sum_{\substack{1\leq l \leq K \\ l \neq k}} \xi_{kl} \mathcal{V}_n[g_k(\vesup{X}{(k)}), g_l(\vesup{X}{(l)})],
    \end{align*}
    using SGD.
    \ENDFOR
    \ENSURE The latent mapping $\{g_k\}_{k=1}^K$.
\end{algorithmic}
\end{algorithm}


\subsection{Modalities selection}
In practical applications, the information contained in different modalities may be redundant. This redundancy necessitates a more rigorous investigation into the modalities used in multi-modal learning frameworks. In this subsection, we elaborate on the modality selection component of DeepSuM.
Selecting the optimal combination of modalities is challenging due to the absence of a standardized method for evaluating their learning utility and NP-hard of subset selection. These difficulties highlight the need for innovative approaches that can efficiently assess and select modalities.


In traditional regression problems, variable selection often hinges on the correlation between predictors and the response variable. For instance, a predictor $X_i$ that is independent of the response $Y$ is generally considered unhelpful for the prediction task and thus excluded from the model.
Following this rationale, we use $\mathcal{V}(g_k(\ve{X}^{(k)}), Y)$ to measure the utility of each modality. If $\mathcal{V}(g_k(\ve{X}^{(k)}), Y) = 0$, the modality $\ve{X}^{(k)}$ is redundant, and thus, not necessary for the task. Conversely, if $\mathcal{V}(g_k(\ve{X}^{(k)}), Y) > 0$, the $k$-th modality should be selected as it contributes positively to predicting $Y$. This quantitative measure assists in systematically evaluating the contribution of each modality, ensuring that only relevant modalities are considered for the final model.
The index sets of active and inactive modalities are defined as
\begin{align*}
\mathcal{A} & =\left\{i \mid \vesup{X}{(i)} \not\perp Y \right\}, \\
\mathcal{I} & =\left\{i \mid \vesup{X}{(i)}   \perp Y \right\} .
\end{align*}

After obtaining the estimate of the utility function $\mathcal{V}_n(g_i(\vesup{X}{(i)}), Y)$, the estimate of the active set is
\begin{align*}
    \hat{\mathcal{A}} = \{i \mid \mathcal{V}_n(g_i(\vesup{X}{(i)}), Y) > \tau_n, \quad i = 1, \dots, K\}
\end{align*}
where $\tau_n$ is a pre-specified constant.

For more general cases, if there are already some modal determinations that would be helpful for downstream tasks, the selection strategy described above still applies, as we consider independence when learning the representation of modalities. We take the scenario involving two modalities for an example. Let $g_1$ represent a sufficient representation of $\ve{X}^{(1)}$, and $g_1$ is independent of $g_2$. If $g_2(\ve{X}^{(2)})$ and $Y$ are independent, then $g_1(\ve{X}^{(1)})$ fully captures the information provided by both modalities $\ve{X}^{(1)}$ and $\ve{X}^{(2)}$. Consequently, introducing $g_2(\ve{X}^{(2)})$ does not enhance the prediction capability.

For simplify, we assume the modality $\ve{X}^{(1)}, \dots, \ve{X}^{(k_0)}$, $k_0 < K$ is already selected, we perform modality selection on $\ve{X}^{(k_0+1)},  \dots, \ve{X}^{(K)}$. The selection procedure is based on the utility function $\mathcal{V}(g_k(\vesup{X}{(k)}), Y), k = k_0 + 1, \dots, K$. The index sets of active and inactive modalities becomes
\begin{align*}
\mathcal{A} & =\left\{i \mid \vesup{X}{(i)} \not\perp Y \mid \ve{X}^{(1)}, \dots, \ve{X}^{(k_0)}\right\}, \\
\mathcal{I} & =\left\{i \mid \vesup{X}{(i)}   \perp Y \mid  \ve{X}^{(1)}, \dots, \ve{X}^{(k_0)}\right\} .
\end{align*}
After obtaining the estimate of the utility function $\mathcal{V}_n(g_i(\vesup{X}{(i)}), Y)$, the estimate of the active set is
\begin{align*}
    \hat{\mathcal{A}} = \{i \mid \mathcal{V}_n(g_i(\vesup{X}{(i)}), Y) > \tau_n, \quad i = k_0+1, \dots, K\}
\end{align*}
where $\tau_n$ is a pre-specified constant. 


In summary, our modality selection measures the contribution of the latent variable corresponding to the modality to $Y$ by using the independence measure. Due to the independence assumption, we do not need to solve the NP-hard problem and only need to compare the $\mathcal{V}_n(g_i(\vesup{X}{(i)}), Y)$ of $K$ modalities.

\section{Theoretical Properties}
In this section, we will establish the theoretical properties of DeepSuM.


\subsection{Convergence of DeepSuM Algorithm}
We first give the relationship between the loss function $F$ and the optimization problem~\ref{opt0}.
\begin{theorem}
    We have $g^{*} \in \operatorname{argmin}_{g \in \mathcal{G}} F(g)$ provided \ref{opt0} holds.
    \label{thm::opt1}
\end{theorem}

According to Theorem~\ref{thm::opt1}, we can estimate $g^{*}$ by solving the empirical version of the objective function $F$ with the sample $\{(\vesub{X}{i}, Y_i), i = 1, 2, \dots, n\}$.

Next, we move to consider the consistency of DeepSuM, and show that the excess risk $F(\hat{g}) - F(g^*)$ converges to zero, where $\hat{g}$ is obtained by $\operatorname{argmin}_{g \in \mathcal{G}} \hat{F}(g)$
and
$$
\hat{F}(g) = -\sum_{k=1}^K \big[\mathcal{V}_n[g_k(\vesup{X}{(k)}), Y]  +\lambda_k \hat{\mathbb{D}}(\mu(g_k(\vesup{X}{(k)})), \gamma_{d_k}) \big] + \sum_{1 \leq k < l \leq K} \xi_{kl}\mathcal{V}_n[g_k(\vesup{X}{(k)}), g_l(\vesup{X}{(l)})].
$$
The latent representation $\hat{g}$ is estimated by feedforward neural networks (FNN). In out algorithm, we use two networks: the representer network $g_{\theta}$ with parameter $\theta$ for estimating $g^{*}$ and discriminator network $D_{\phi}$ with parameter $\phi$ for estimating the discriminator $D$. To obtain the convergence of the estimate latent representation, we first introduce some assumptions on the structure of the networks.

The representation network $g_{\theta}$ contains $K$ networks for each modality, let $\mathbf{G}_{\mathcal{H}, \mathcal{W}, \mathcal{S}} = \mathbf{G}_{\mathcal{H}_1, \mathcal{W}_1, \mathcal{S}_1}^1 \oplus \cdots \oplus \mathbf{G}_{\mathcal{H}_K, \mathcal{W}_K, \mathcal{S}_K}^K$ be the set of the ReLU networks with parameter $\theta$, depth $\mathcal{H}$, width $\mathcal{W}$, size $\mathcal{S}$. 
Here, $\mathbf{G}_{\mathcal{H}_k, \mathcal{W}_k, \mathcal{S}_k}^k$ is the network for $k$-th modality, the depth $\mathcal{H}_k$ refers to the number of hidden layers, so the network has $\mathcal{H}_k+1$ layers in total. A $(\mathcal{H}_k+1)$-vector $\left(w_{0k}, w_{1k}, \ldots, w_{\mathcal{H}_k k}\right)$ specifies the width of each layer, where $w_{0k}$ is the dimension of the input data and $w_{\mathcal{H}_k k}=d_k$ is the dimension of the output. The width $\mathcal{W}_k=\max \left\{w_{1k}, \ldots, w_{\mathcal{H}_k k}\right\}$ is the maximum width of the hidden layers. The size $\mathcal{S}_k=\sum_{i=0}^{\mathcal{H}_k}\left[w_{ik} \times\left(w_{(i+1) k}\right)\right]$ is the total number of parameters in the network. Similarly, we define the discriminator network $D_\phi$ as $\mathbf{D}_{\tilde{\mathcal{H}}, \tilde{\mathcal{W}}, \tilde{\mathcal{S}}} = \mathbf{D}_{\tilde{\mathcal{H}}_1, \tilde{\mathcal{W}}_1, \tilde{\mathcal{S}}_1}^1 \oplus \cdots \oplus \mathbf{D}_{\tilde{\mathcal{H}}_K, \tilde{\mathcal{W}}_K, \tilde{\mathcal{S}}_K}^K$ with parameter $\phi$ depth $\tilde{\mathcal{H}}$, width $\tilde{\mathcal{W}}$, size $\tilde{\mathcal { S }}$.

The latent representation obtained by DeepSuM algorithm above is 
$$\hat{g} = \operatorname{argmin}_{g \in \mathbf{G}_{\mathcal{H}, \mathcal{W}, \mathcal{S}}} \hat{F}(g).$$
All three parts in $\hat{g}$ are unbiased and consistent estimates. To obtain the convergence of $hat{g}$, we can analyse the excess risk $F(\hat{g}) - F(g^*)$. We further introduce some assumptions on the parameter and the model.
\begin{assumption}
    The target representation $g^*_k$ is Lipschitz continuous with Lipschitz constant $L_{1k}$ for $k = 1, \dots, K$, and $\mathbb{E}[\|g^*_k\|]$
    \label{ass1}
\end{assumption}
\begin{assumption}
    For every $g_k \in  \mathcal{G}_{\mathcal{H0}_k, \mathcal{W}_k, \mathcal{S}_k}$, we assume the density ratio $r_k(z)=\frac{\mathrm{d} \mu(g_k)}{\mathrm{d} \gamma_{d_k}}(z)$ to be Lipschitz continuous with Lipschitz constant $L_{2k}$, and $c_{1k} \leq r_k(z) \leq c_{2k}$ for some constants $0<c_{1k} \leq c_{2k}<\infty$, $k = 1, \dots, K$. 
    \label{ass2}
\end{assumption}

\begin{assumption}
    $\operatorname{supp}\left(\mu(\vesup{X}{(k)})\right)$ is contained in a compact set, say $\left[-B_{1k}, B_{1k}\right]^{p_k}$ with a finite $B_{1k}$ and denote its density function as $f_k(x)$. $Y$ is bounded almost surely, say $\|Y\| \leq C_0$ a.s..
    \label{ass3}
\end{assumption}
For the loss function $F(g)$, we turn it into a function of the latent representation  $g_k$ of each modality, so we introduce Assumption on $g_k$. More specifically, Assumption~\ref{ass1},\ref{ass2} are commonly used regularity and smooth assumptions in non-parametric regression, while Assumption~\ref{ass3}  makes restrictions on the support set and density function of each modality.

Next, we make restrictions on the network parameter as follows:
\begin{assumption}
    The parameter of the representation network satisfies:  $\mathcal{H}_k=O(\log n)$, $\mathcal{W}_k=O\left(n^{\frac{p_k}{2(2+p_k)}} / \log n\right)$,  $\mathcal{S}_k=O\left(d_k  n^{\frac{p_k}{2+p_k}} / \log ^4(n p_k d_k)\right)$, and $\|g_k\|_{L^{\infty}} \leq 2\left\|g_k^*\right\|_{L^{\infty}}, \forall g_k \in \mathbf{G}_{\mathcal{H}_k, \mathcal{W}_k, \mathcal{S}_k}^k$ for $k = 1, \dots, K$.
    \label{ass4}
\end{assumption}

\begin{assumption}
    The parameter of the discriminator network satisfies:  $\tilde{\mathcal{H}}_k=O(\log n)$, $\tilde{\mathcal{W}}_k=O\left(n^{\frac{d_k}{2(2+d_k)}} / \log n\right)$, size $\tilde{\mathcal{S}}=$ $O\left(n^{\frac{d_k}{2+d_k}} / \log ^4(n p_k d)\right)$, and $\|D_k\|_{L^{\infty}} \leq 2 B_{2k}, \forall D \in \mathbf{D}_{\tilde{\mathcal{H}}_k, \tilde{\mathcal{W}}_k, \tilde{\mathcal{S}}_k}^k$, $B_{2k}=\max \left\{\log c_{1k}, \log c_{2k}\right\}+1$ for $k = 1, \dots, K$.
    \label{ass5}
\end{assumption}


Now we can obtain the consistency of the estimated representation map.

\begin{theorem}
    Suppose Assumption~\ref{ass1}-\ref{ass5} hold and $\lambda_k = O(1), k = 1, \dots, K$, $\xi_{kl} = O(1), 1 \leq k < l \leq K$, then we have
    $$E[F(\hat{g}) - F(g^*)] \rightarrow 0.$$
    \label{thm::con1}
\end{theorem}

Theorem~\ref{thm::con1} gives the convergence of the risk function, indicating that the sufficient representation estimated by DeepSuM is consistent, ensuring the accuracy of learning multimodal representations.

\subsection{Modality Selection Properties}
Now, we turn to analyse the selection properties of our proposed method. 
To verify the asymptotic properties of the selection procedure, we introduce the following assumption:
\begin{assumption}
    The utility functions satisfy

    (i) Both $g_i(\vesup{X}{(i)})$ and $Y$ satisfy the sub-exponential tail probability uniformly in $K$. That is, there exists a positive constant $s_0$ such that for all $0<s \leq 2 s_0$,
$$
\sup_p \max _{1 \leq i \leq K} E\left\{\exp \left(s\left\|g_i(\vesup{X}{(i)})\right\|\right)\right\}<\infty \text {, and } E\left\{\exp \left(s\|Y\|\right)\right\}<\infty \text {. }
$$
    
    (ii) there exist some constants $c>0$ and $0 \leq \kappa<1 / 2$ such that $\min_{i \in \mathcal{A}} \mathcal{V}(g_i(\vesup{X}{(i)}), Y) \geq 2 c n^{-\kappa}$.

    (iii) assume $\log (K)=o\left(n^{1-2 \kappa}\right)$, where $\kappa$ is defined in condition (ii).
    \label{ass::sel}
\end{assumption}

Condition 1 follows immediately when latent representations and $Y$ are bounded uniformly. Condition 2 ensures that the strength of the modal signal is not too small to be detected. Condition 3 places a constraint on the number of modalities, but in most cases, the number of modalities is finite, so the condition is naturally established.

\begin{theorem}[Strong selection consistency]
 If Assumption~\ref{ass::sel} holds, there exists a positive constant $C > 0$ such that
 $$
P\left(\left|\mathcal{V}_n[Z, Y]-\mathcal{V}[Z, Y]\right| \geq 4 \varepsilon\right) \leq 2 \exp \left(-\varepsilon^2 n^{1-2 \gamma}\right)+2 n C \exp \left(-t n^\gamma / 8\right)
$$
hold for $t > 0$. Moreover, we have
$$
\mathbb{P}\left(\mathcal{A}=\hat{\mathcal{A}}\right) \rightarrow 1 \text { as } n \rightarrow \infty .
$$
    \label{thm::selpro}
\end{theorem}

Theorem~\ref{thm::selpro} shows the consistency of our method in modality selection. It can accurately select the modalities with signals and discard irrelevant modalities, ensuring high efficiency in the application process.

\subsection{Linear Case}
In this subsection, we focus on a composite linear data generating model to theoretically verify the properties of our multimodal learning framework and modality selection procedure.

Formally, we assume the data is generated by
$$
Y= \ve{X} \mathbf{A} \boldsymbol{\beta}^{\star}  +\epsilon
$$
Here $\ve{X} = (X^{(1)}, X^{(2)},  \dots, X^{(K)}) \in \mathbf{R}^{d_1 + \dots + d_K}$, which consists of $K$ modalities, $d = d_1 + \dots + d_k$, $\mathbf{A} \in \mathbf{R}^{d \times r}$ is the latent matrix with dimension $r$, $\epsilon$ is independent of $\ve{X}$ and has zero-mean and bounded second moment $\sigma^2$. 
The function class of the latent mapping and target mapping are defined as
$$
\begin{aligned}
& \mathcal{G}=\left\{g \mid g(\ve{X})=\ve{X}\mathbf{A} , \mathbf{A} \in \mathbb{R}^{d \times r}\right\} \\
& \mathcal{H}=\left\{h \mid h(\mathbf{z})= \mathbf{z} \boldsymbol{\beta} , \boldsymbol{\beta} \in \mathbb{R}^r,\|\boldsymbol{\beta}\| \leq C\right\}
\end{aligned}
$$
where $C$ is a constant. The loss is chosen as $l_2$ loss, then we have 
$$
\eta(g) = \eta(\mathbf{A}) =
\mathbb{E}\left\{\ell(h \circ g(\ve{X}), y)-\ell\left(h^{\star} \circ g^{\star}(\ve{X}), y\right)\right\}=\inf _{\boldsymbol{\beta}:\|\boldsymbol{\beta}\| \leq C_b} \mathbb{E}\left[\left|\ve{X} \mathbf{A} \boldsymbol{\beta}-  \ve{X} \mathbf{A}^{\star }\boldsymbol{\beta}^{\star}\right|^2\right]
$$

Let's first consider the properties of the utility function under the assumption of a linear model, and we get the following theorem:
 
\begin{theorem}
Assume the latent matrix based on the proposed algorithm is $\mathbf{A}_1$, $\mathbf{A}_2$ is another latent matrix satisfied $\eta(\mathbf{A}_1) = \eta(\mathbf{A}_2)$ for all $ \boldsymbol{\beta}^{\star}$, 
then we have $\mathbf{A}_2 = \mathbf{A}_1 \mathbf{R}$, where $\mathbf{R}$ is a orthogonal matrix.
    \label{thm::linearorth}
\end{theorem}

Theorem~\ref{thm::linearorth} asserts that if two latent matrices share the same utility function, they must differ by exactly one orthogonal matrix. In the development of the model described in Section 2, we adopted sufficient dimensional reduction as the criterion for learning the representations of latent matrices. Additionally, we ensured that the conditional independence of these matrices is preserved even when subjected to transformations by an invertible matrix. This preservation aligns with the theorem regarding orthogonal matrices.

In our multimodal learning algorithm, we mandate that the latent representations learned by each modality remain independent. This independence is crucial not only for facilitating the analysis of our downstream tasks but also for establishing a theoretical foundation. Specifically, we will demonstrate that this approach results in a reduced variance for the learned parameter $\boldsymbol{\beta}$.

\begin{theorem}
    Assume we have already learned the latent representation of $K$ modalities $\ve{X}$ with $\mathbf{A}$, $\vesub{Z}{0} =  \ve{X} \mathbf{A}$, and now have a newly introduced modality $X^{(K+1)}$ with a latent representation $\vesub{Z}{1}$ by our method. For another latent representation $\vesub{Z}{2}$, $\vesub{Z}{2}= \vesub{Z}{1} R$, $R$ is a transition matrix , the estimates of $\boldsymbol{\beta}_1$ and $\boldsymbol{\beta}_2$ satisfy $\var(\hat{\boldsymbol{\beta}_1}) \leq \var(\hat{\boldsymbol{\beta}_2})$.
    \label{thm::linearvar}
\end{theorem}

Theorem~\ref{thm::linearvar} illustrates that our method has lower variance in downstream tasks than its other latent representations, thanks to the fact that the latent representations between each modality are independent. Moreover, based on the result in \citep{huang2021makes}, we know if the new $X^{(K+1)}$ is not related to $Y$, then the utility function $\eta$ will not increase.

\section{Numerical Studies}\label{method}

In this section, we will conduct numerical simulation experiments to validate the effectiveness of our proposed method. Our focus is on additional modalities in the presence of pre-existing ones. In practical scenarios, some modalities are readily accessible, while others may be costly or challenging to acquire. Determining the necessity of integrating these additional modalities into the analysis is crucial.

We will specifically explore three cases:
(i) The original modality encompasses all the necessary information about the response variable and the additional modalities might either provide complete or partial information.
(ii) The original modality contains only partial information, while the additional modality offers complementary or the same information.
(iii) The original modality contains only partial information, but with the additional modality presenting complementary information at differing signal strengths.
Through these simulations, we aim to assess the impact of integrating various modalities on the robustness and accuracy of our analytical method.

For the latent variable $Z$, we generate it by standard normal distribution with dimension $d = 3$. The response variable is generated by a nonlinear function of the latent variable. We mainly consider three cases 
\begin{itemize}
    \item Scenario 1: $Y = (Z_1 + Z_2)^2 + (1 + \exp(X_2))^2 + \epsilon$;
    \item Scenario 2: $Y = \sin(\frac{\pi}{10}(Z_1 + Z_2)) + Z_2^2 + \epsilon$;
    \item Scenario 3: $Y = \sqrt{Z_1^2 + Z_2^2} \log(\sqrt{Z_1^2 + Z_2^2}) + \epsilon$;
\end{itemize}
where $\epsilon$ is i.i.d. noise followed $N(0, \sigma^2)$.

We first assume that the predictor variable modality $X \in \mathbf{R}^p$ is already obtained and is generated as follows:
\begin{align*}
    X = Z A_x + \epsilon_x,
\end{align*}
where $A_x \in \mathbf{R}^{d  \times p}$ is the transition matrix and $\epsilon_x$ is standard normal noise. For the additional modalities, we consider $ U, V, W \in \mathbf{R}^q$, which are generated by
\begin{align*}
    U = Z A_u + \epsilon_u, \quad V = Z A_v + \epsilon_v, \quad W = Z A_w + \epsilon_w,
\end{align*}
where $A_u, A_v, A_w \in \mathbf{R}^{r  \times q}$ are the transition matrices of additional modalities, and $\epsilon_u, \epsilon_v, \epsilon_w$ are normal noise with variance $\sigma_u^2, \sigma_v^2 , \sigma_w^2 = 1$. All the transition matrices are generated by standard normal distribution.

For three cases mentioned above, we design the simulations as follow:
\begin{itemize}
    \item Case 1: The first row of $A_u$ and the second row of $A_v$ are set to 0, $A_w = 0$.
    \item Case 2: The second row of $A_x, A_v$ are set to 0, the first row of $A_u$ is set to 0, $A_w = 0$.
    \item Case 3: The second row of $A_x$ are set to 0, the first row of $A_u, A_v, A_w$ is set to 0, the second row of $A_u, A_v, A_w$ are the same, $\sigma_u^2 = 1, \sigma_v^2 = 2, \sigma_w^2 = 4$.
\end{itemize}
From the configurations in the three scenarios, it is evident that the information predominantly resides in the first two dimensions of $Z$. In the first case, no modifications were made to the transition matrix of $X$, indicating that $X$ encapsulates all necessary information. Additionally, the first row of the transformation matrix for $U$ is set to zero, signifying that $U$ lacks information pertaining to $Z_1$. Similarly, $V$ does not contain information about $Z_2$, and $W$ consists solely of noise.
In the second case, $X$ is missing information about $Z_2$, while $U$ does not include information on $Z_1$. Likewise, $V$ is devoid of details about $Z_2$, and $W$ remains purely noise.
In the third case, $X$ is again missing information related to $Z_2$. In this case, $U$, $V$, and $W$ all lack information about $Z_1$, albeit with varying levels of noise.

To obtain the latent representation of each modality, we conduct a three-layer neural network, where each layer has $32$, $16$, and $8$ nodes respectively. The discriminator consists of a two-layer neural network, with $16$ and $8$ nodes in each layer respectively. The dimension of latent space is $5$.
The regression of $Y$ is done by a neural network with two layers, $16$ and $8$ nodes respectively. We use adam as the solver, the activation function is RELU, the maximum number of iterations is $10000$, and the learning rate is $0.001$.

For other variable parameters, we consider sample sizes 
$n = 3000, 5000$, modalities dimensions $(p, q) = (10, 10)$, noise of the response $\sigma^2 = 1$.
Each experiment is repeated $100$ times, and we report the MSE and the most likely selected modality.

\begin{table}[ht]
\centering
\caption{MSE of Case 1 with parameters $p=10$, $q=10$, $\sigma=1$.}
\begin{tabular}{|c|l|c|c|c|}
\hline
& & \textbf{Scenario 1} & \textbf{Scenario 2} & \textbf{Scenario 3} \\
\hline
\multirow{4}{*}{$n = 3000$} &\textbf{MSEX}     &  1.51 (0.921) & 0.438 (0.32) & 0.281 (0.13) \\
& \textbf{MSEXU}     & 1.915 (0.843) & 0.458 (0.148) & 0.382 (0.098) \\
& \textbf{MSEXV}   & 2.234 (1.35) & 0.66 (0.458) & 0.38 (0.096)   \\
& \textbf{MSEXW}  & 2.602 (1.438) & 0.714 (0.484) & 0.438 (0.158) \\
\hline
\multirow{4}{*}{$n = 5000$} &\textbf{MSEX}     &  1.907 (1.305) & 0.443 (0.141) & 0.294 (0.07)\\
& \textbf{MSEXU}     & 2.114 (1.209) & 0.436 (0.064) & 0.361 (0.058)\\ 
& \textbf{MSEXV}   & 2.363 (1.52) & 0.571 (0.169) & 0.359 (0.052)   \\
& \textbf{MSEXW}  & 2.777 (1.773) & 0.619 (0.188) & 0.4 (0.072) \\
\hline
\end{tabular}
\label{sim::case1}
\end{table}

\begin{table}[ht]
\centering
\caption{MSE of Case 2 with parameters $p=10$, $q=10$, $\sigma=1$.}
\begin{tabular}{|c|l|c|c|c|}
\hline
& & \textbf{Scenario 1} & \textbf{Scenario 2} & \textbf{Scenario 3} \\
\hline
\multirow{4}{*}{$n = 3000$} &\textbf{MSEX}     &  10.435 (5.91) & 1.718 (0.877) & 0.514 (0.207) \\
& \textbf{MSEXU}     & 7.078 (2.035) & 0.62 (0.175) & 0.459 (0.094)	 \\
& \textbf{MSEXV}   & 16.009 (4.729) & 2.679 (0.642) & 0.727 (0.133)	 \\
& \textbf{MSEXW}  &  17.85 (5.543) & 2.861 (0.684) & 0.829 (0.143)\	 \\
\hline
\multirow{4}{*}{$n = 5000$} &\textbf{MSEX} &   11.283 (4.513) & 1.845 (0.752) & 0.556 (0.136)\\
& \textbf{MSEXU}     & 6.492 (1.632) & 0.607 (0.227) & 0.404 (0.055)\\
& \textbf{MSEXV}   & 15.877 (3.382) & 2.54 (0.545) & 0.724 (0.096)\\
& \textbf{MSEXW}  & 17.265 (3.781) & 2.768 (0.543) & 0.792 (0.11)\\
\hline
\end{tabular}
\label{sim::case2}
\end{table}

\begin{table}[ht]
\centering
\caption{MSE of Case 3 with parameters $p=10$, $q=10$, $\sigma=1$.}
\begin{tabular}{|c|l|c|c|c|}
\hline
& & \textbf{Scenario 1} & \textbf{Scenario 2} & \textbf{Scenario 3} \\
\hline
\multirow{4}{*}{$n = 3000$} &\textbf{MSEX}     & 10.273 (7.881) & 1.634 (1.159) & 0.538 (0.27)	 \\
& \textbf{MSEXU}     & 7.617 (3.061) & 0.695 (0.281) & 0.526 (0.126) \\
& \textbf{MSEXV}   & 9.789 (3.365) & 1.136 (0.452) & 0.624 (0.128) \\
& \textbf{MSEXW}  &  13.506 (4.127) & 1.913 (0.593) & 0.769 (0.142) \\
\hline
\multirow{4}{*}{$n = 5000$} &\textbf{MSEX} &    10.937 (6.158) & 1.771 (0.885) & 0.596 (0.19) \\
& \textbf{MSEXU}     & 6.871 (2.148) & 0.608 (0.265) & 0.478 (0.088) \\
& \textbf{MSEXV}   & 9.241 (2.776) & 1.012 (0.321) & 0.57 (0.108) \\
& \textbf{MSEXW}  & 12.863 (3.533) & 1.759 (0.479) & 0.701 (0.127) \\
\hline
\end{tabular}
\label{sim::case3}
\end{table}

\begin{table}[ht]
\centering
\caption{Selection proposition with parameters $p=10$, $q=8$, $\sigma=1$.}
\begin{tabular}{|cc|ccc|ccc|ccc|}
\hline
\multicolumn{2}{|c|}{\multirow{2}{*}{}} & \multicolumn{3}{c|}{\textbf{Scenario 1}} & \multicolumn{3}{c|}{\textbf{Scenario 2}} & \multicolumn{3}{c|}{\textbf{Scenario 3}}  \\
& & \textbf{U} & \textbf{V} & \textbf{W} & \textbf{U} & \textbf{V} & \textbf{W}  & \textbf{U} & \textbf{V} & \textbf{W} \\
\hline
\multirow{2}{*}{Case 1 } & $n = 3000$ 
 & 99.6\% & 0.4\% & 0\% & 99.6\% & 0.4\% & 0\% & 46.4\% & 53.6\% & 0\% \\
& $n = 5000$ &  100\% & 0\% & 0\% & 99.6\% & 0.4\% & 0\% & 48.8\% & 51.2\% & 0\%\\
\hline
\multirow{2}{*}{Case 2 } &  $n = 3000$ & 99.6\% & 0.4\% & 0\% & 100\% & 0\% & 0\% & 96.4\% & 3.6\% & 0\% \\
& $n = 5000$ & 98.8\% & 1.2\% & 0\% & 98\% & 1.6\% & 0.4\% & 98.4\% & 1.6\% & 0\%\\
\hline
\multirow{2}{*}{Case 3 } & $n = 3000$ & 96.4\% & 3.6\% & 0\% & 98.4\% & 1.6\% & 0\% & 96.4\% & 3.6\% & 0\% \\
& $n = 5000$ & 99.6\% & 0.4\% & 0\% & 98.8\% & 1.2\% & 0\% & 98\% & 2\% & 0\%\\
\hline
\end{tabular}
\end{table}

As demonstrated in Table~\ref{sim::case1}, if a newly added modality contributes no additional signal, our method identifies it as irrelevant noise. Consequently, this does not enhance the performance of downstream tasks and even degrade performance. This underscores the importance of excluding irrelevant modalities from downstream task analyses.

In Table~\ref{sim::case2}, the original modality $X$ is deficient in crucial information $Z_1$. Introducing complementary information via modality $U$ markedly enhances downstream task performance. Conversely, the introduction of modality $V, W$, which only adds repetitive information or noise, does not improve performance. Moreover, as Table~\ref{sim::case3} shows, introducing complementary information with varying signal strengths reveals that stronger noise levels can obliterate the original signal, thereby degrading the performance of downstream tasks to levels even below those achieved by using $X$ alone.

In general, our method can learn the representation of additional modalities well and extract useful signals given a modality. For modalities with repeated information, our method learns their potential representation as irrelevant normal noise, and its addition may lead to poor performance of downstream tasks. Therefore, in the problem of multimodal learning, we must screen additional modalities and introduce truly valuable variables, which can not only improve the performance of downstream tasks, but also improve computational efficiency.

\section{Real Data Analysis}
\subsection{Kidney cortex cells classification}
In this study, we explore the BBBC05128 dataset \citep{woloshuk2021situ}, which is part of the Broad Bioimage Benchmark Collection \citep{ljosa2012annotated}. The dataset comprises $236,386$ human kidney cortex cells segmented from $3$ reference tissue specimens and categorized into $8$ groups. Each gray-scale image has dimensions of $32 \times 32 \times 7$ pixels, where $7$ represents the number of cell slices.
Our analysis focuses on classifying four types of cell: glomerular, peritubular, podocytes, thick ascending limb, with $8,256; 21,167; 10,802; 34,970$ samples respectively. We treat the $7$ slices as distinct modalities and select those that are most informative for classification.

We implemented a convolutional neural network (CNN) with layers containing $16, 64, 64$ channels respectively. Pooling layers are positioned at the second and third layers. The network converges into a fully connected layer, and the latent space dimension is set to $16$. 

\begin{figure}[h!]
	\centerline{\includegraphics[width = 0.8\textwidth, height = 0.8\textwidth]{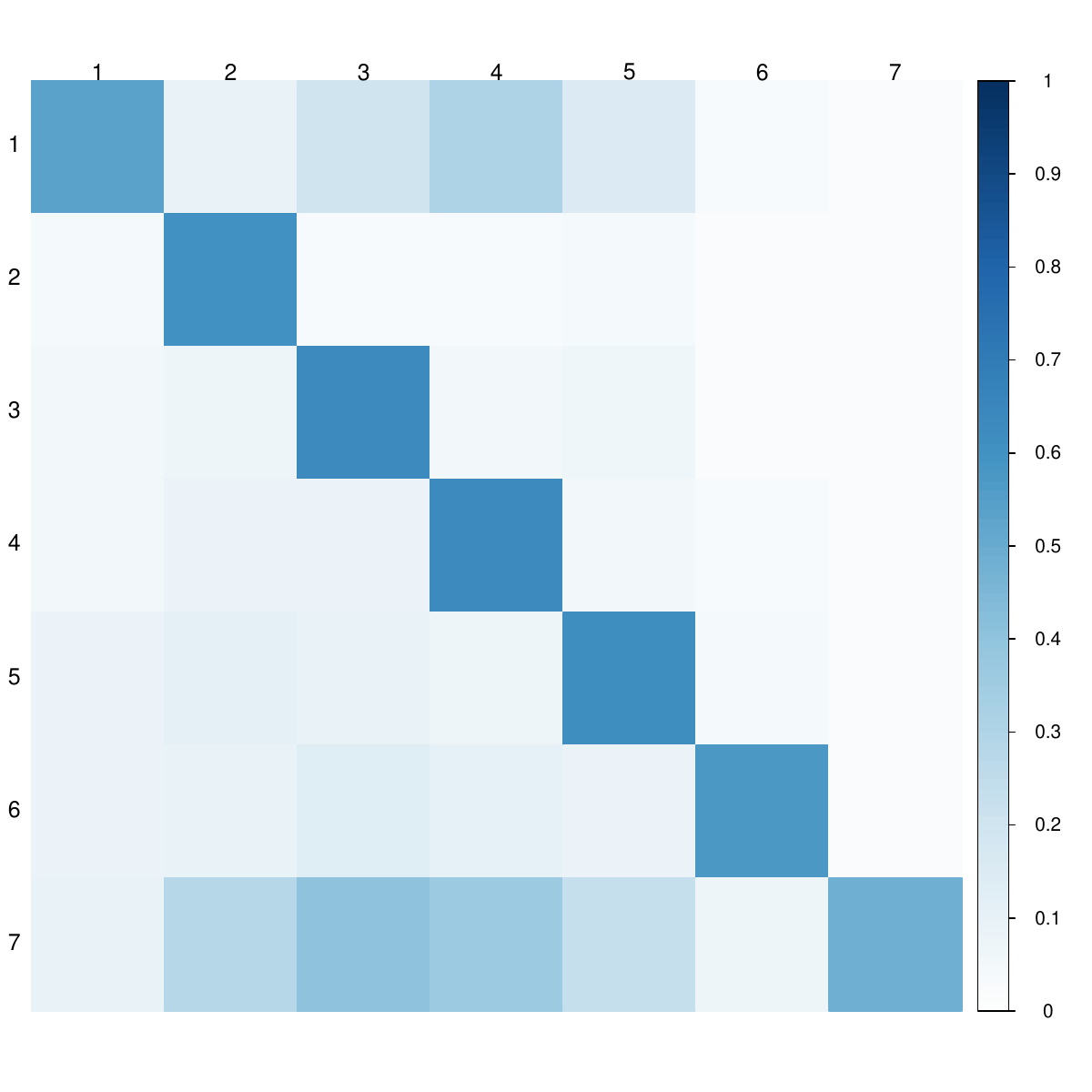}}
	\caption{Distance correlation in cell classification}
	\label{fig_heatmap}
\end{figure}

Following this, a single-layer network with $16$ nodes is used for classification. We divided the source dataset into training, validation, and test sets with a ratio of $6:2:2$, the batch size is $128$.
To investigate the interplay between different modalities and their impact on classification accuracy, we initially select one slice and then sequentially integrate each of the remaining six slices. The efficacy of this approach was evaluated through $100$ experimental replicates, focusing on distance correlation and accuracy. The results are presented in Table~\ref{real0::acc} and Figure~\ref{fig_heatmap}. The diagonal entries reflect scenarios where only one slice is utilized, while off-diagonal entries indicate that the slice corresponding to the row was selected first, followed by the slice corresponding to the column. The distance correlation values in the non-diagonal parts thus represent the correlation after the slice in the row has been selected.

\begin{table}[h]
\centering
\caption{Accuracy in cell classification, the highest accuracy is indicated in bold.}
\begin{tabular}{|c|ccccccc|}
\hline
 & 1 & 2 & 3 & 4 & 5 & 6 & 7 \\ \hline
1 & 0.689 & 0.725 & 0.753 & \textbf{0.773} & 0.741 & 0.706 & 0.695 \\ \hline
2 & 0.736 & 0.728 & \textbf{0.737} & 0.737 & 0.740 & 0.736 & 0.735 \\ \hline
3 & 0.750 & 0.753 & 0.740 & 0.749 & \textbf{0.756} & 0.745 & 0.744 \\ \hline
4 & 0.751 & 0.759 & \textbf{0.761} & 0.741 & 0.749 & 0.746 & 0.745 \\ \hline
5 & 0.755 & \textbf{0.766} & 0.757 & 0.748 & 0.732 & 0.737 & 0.736 \\ \hline
6 & 0.727 & 0.736 & \textbf{0.753} & 0.745 & 0.734 & 0.704 & 0.707 \\ \hline
7 & 0.684 & 0.737 & \textbf{0.769} & 0.758 & 0.727 & 0.691 & 0.654 \\ \hline
\end{tabular}
\label{real0::acc}
\end{table}


From Table~\ref{real0::acc}, it is evident that the accuracy across each row is quite close, especially the 2nd to 5th slices. In contrast, the first, sixth, and seventh slices contain less information, and their prediction accuracy is not high when used alone. After introducing any of the four middle slices, the prediction accuracy is significantly improved. 
For the five middle slices, the correlations introduced into other slices is very small. In contrast, for the first and seventh slices, introducing any of the second to fifth slices has a larger correlation coefficient. This indicates that the second to fifth slices contain the most information for classification, while the first and seventh slices have incomplete information and need additional supplementation from the four slices above.
Analyzing the performance across all seven slices, slices 3 and 4 emerge as having the most effective predictive capability.

\subsection{Breast invasive carcinoma survival status analysis}
In this study, we analyzed data on breast invasive carcinoma (BRCA) sourced from the Cancer Genome Atlas Program (TCGA). This comprehensive dataset includes various omics profiles and clinical data from over 10,000 cancer patients, providing a significant resource for bioinformatics research.
For our analysis, the primary outcome variable was the survival status of BRCA patients. We categorized survival based on whether the time exceeded two years, dividing the cohort into two groups for classification purposes. For predictors, we considered three types of data: protein, RNA (exon expression), and DNA (copy number variation, CNV).
From the available data, we selected a subset where all three types of data intersected, resulting in a total of $738$ samples. Of these, $389$ patients survived longer than two years, while $349$ did not meet this threshold.

Protein expression was analyzed using a reverse phase protein array (RPPA) with $281$ identifiers. Exon-level transcription estimates were quantified in RPKM values (Reads Per Kilobase of exon model per Million mapped reads) across $239,322$ identifiers. Copy number variations (CNVs) were estimated using the GISTIC2 method and categorized into five levels: $-2$, $-1$, $0$, $1$, $2$, representing homozygous deletion, single copy deletion, diploid normal copy, low-level copy number amplification, and high-level copy number amplification, respectively. The dataset for CNV analysis included $24,776$ identifiers.

We removed the identifiers that were missing from all the samples, and the protein had $166$ identifiers. For the exon data, gene names in the RNA-seq expression matrix were converted from ENSEMBL IDs to gene symbols. Expressions from ENSEMBL IDs sharing the same gene symbol were averaged. Genes with total expression levels below a predetermined threshold across all samples were excluded from further analysis. Using the R package edgeR \citep{Robinson2009edgeR}, we selected $262$ genes associated with survival status based on an FDR of less than $0.05$ and a logarithm of fold change greater than $1$. For the CNV data, we employed a chi-square test for each gene against the survival state and performed a Benjamini-Hochberg (BH) correction \citep{benjamini1995controlling}, resulting in the selection of $358$ genes with an FDR less than $0.25$.

For the protein data, we conduct a two-layer neural network, where each layer has $128$ and $16$ nodes respectively, a $0.5$ dropout layer is added to the first layer. For the RNA data, we conduct a two-layer neural network, where each layer has $32$ and $16$ nodes respectively, a $0.5$ dropout layer is added to the first layer. For the DNA data, we conduct a two-layer neural network, where each layer has $32$ and $16$ nodes respectively. Each data modality is represented in a latent space with a dimensionality of $8$. After learning these latent representations, we utilized XGBoost \citep{chen2016xgboost} to predict the survival status of the patients. The configuration for XGBoost included a maximum tree depth of $3$, with a total of $100$ trees, and a learning rate of $0.3$.
To explore the interplay between the different modalities and their impact on survival status prediction, we devised a method where we initially select one modality and subsequently integrate the remaining two. The dataset was split into training, validation, and test sets in a $6:2:2$ ratio. We assessed the effectiveness of our approach by reporting the Distance correlation and AUC under $500$ replicates of the experiment. The results are detailed in Table~\ref{real::dc} and Table~\ref{real::auc}.

\begin{table}[ht]
\centering
\caption{Distance correlation in BCRA analysis.}
\begin{tabular}{|c|c|c|}
\hline
\textbf{First Component} & \textbf{Second Component} & \textbf{Third Component} \\
\hline
\multirow{2}{*}{Protein (0.5071)} & RNA (0.0768)     &  DNA (0.0911) \\
\cline{2-3} 
&  DNA (0.0871)    & RNA (0.0816)  \\
\hline
\multirow{2}{*}{RNA (0.3242)} & Protein (0.4207)     &  DNA (0.0633) \\
\cline{2-3} 
&  DNA (0.1251)    & Protein (0.4711)  \\
\hline
\multirow{2}{*}{DNA (0.2605)} & Protein (0.4562)     &  RNA (0.0555) \\
\cline{2-3} 
&  RNA (0.2006)    & Protein (0.3995)  \\
\hline
\end{tabular}
\label{real::dc}
\end{table}

\begin{table}[ht]
\centering
\caption{AUC in BCRA analysis.}
\begin{tabular}{|c|c|c|}
\hline
\textbf{First Component} & \textbf{Second Component} & \textbf{Third Component} \\
\hline
\multirow{2}{*}{Protein (0.5988)} & RNA (0.5943)     &  DNA (0.6017) \\
\cline{2-3} 
&  DNA (0.6082)    & RNA (0.6044)  \\
\hline
\multirow{2}{*}{RNA (0.5334)} & Protein (0.5793)     &  DNA (0.5858) \\
\cline{2-3} 
&  DNA (0.5597)    & Protein (0.5901)  \\
\hline
\multirow{2}{*}{DNA (0.5688)} & Protein (0.6047)     &  RNA (0.6004) \\
\cline{2-3} 
&  RNA (0.5675)    & Protein (0.5957)  \\
\hline
\end{tabular}
\label{real::auc}
\end{table}

According to the results summarized in Table~\ref{real::dc} and Table~\ref{real::auc}, the protein modality exhibits the highest distance correlation and AUC values with regard to survival status when used alone. It is followed by DNA, while RNA demonstrates the lowest performance. When the protein modality is selected initially, adding either DNA or RNA does not significantly enhance predictive performance; however, DNA contains more supplementary information than RNA. Conversely, if either DNA or RNA is chosen first, it becomes essential to include the remaining two modalities for optimal results, with protein being the preferred choice as the second modality. Moreover, if the first two selected modalities are RNA and DNA, then introducing the protein modality as the third is necessary.

In summary, if only one modality can be selected for data collection, protein is unequivocally the superior choice due to its strong correlation with survival status. Should there be an opportunity to collect data from an additional modality, DNA should be prioritized. Generally, once protein data is available, RNA provides minimal additional value for analyzing survival status, whereas DNA offers some unique insights not captured by protein data alone.

Integration of several types of patient data in a computational framework can accelerate the identification of more reliable biomarkers, especially for prognostic purposes. \cite{zhang2016network} showed that mRNA expression and DNA methylation features provided the highest contribution to the detection of patient survival, followed by CNV and miRNA features among four cancer types lung squamous cell carcinoma (LUSC), glioblastoma multiforme (GBM), kidney renal clear cell carcinoma (KIRC) and ovarian serous cystadenocarcinoma (OV). RNA-Seq data has better performance compared with proteomic (RPPA) data in survival time prediction for KIRC, GBM, LUSC \citep{Zerrin2017}. TCGA used a reverse-phase protein array (RPPA) analysis of 172 proteins (including 31 phosphoproteins with phospho-specific antibodies) to generate a signature associated with the risk of tumor recurrence \citep{yang2013predicting}

Overall, this work illustrates the ability of proteomics to com- plement genomics in providing additional insights into pathways and processes that drive ovarian cancer biology and how these pathways are altered in correspondence with clinical pheno- types.

In this subsection, our main discovery is that we proposed a method that can comprehensively evaluate the performance of multiple modal data such as protein, RNA, DNA, etc. for predicting survival status, and we can also integrate different modalities together. In the case of BCRA, our method shows that protein data contains the most information, DNA contains a small amount of complementary information, and RNA contains almost no additional information. Although there is still no consensus on which of the three modalities is more helpful for predicting survival status in current research. However, our method can evaluate and select the utility of different modalities to a certain extent, and provide guidance for further research on the mechanism of cancer.


\section{Conclusion and Discussion}\label{conclude}
In this study, we introduce a multimodal learning method that leverages feature representation and sufficient dimension reduction, adaptable across diverse modality spaces. Our approach effectively assesses and selects modalities based on their relevance and contribution to the learning process, enhancing both model efficiency and interpretability.

In terms of extending our research, several avenues are ripe for exploration. First, investigating different types of dependency measures could provide deeper insights into how modalities interact and influence each other, potentially revealing more effective ways to assess modality significance.
Secondly, establishing a theoretical framework to validate the superiority of sufficient representation normality remains a challenge. While our simulation experiments suggest that normality facilitates the accurate estimation of modality independence, a solid theoretical justification is still needed. Further theoretical research could solidify our understanding and provide more robust guidelines for employing normality in multimodal learning.
Lastly, integrating generative models into the process of learning and representing modalities represents a promising direction. This approach could enhance the capability of our framework to handle more complex and diverse datasets, potentially leading to broader applications and deeper understanding of multimodal interactions.







\bibliographystyle{chicago}
\bibliography{mybibtex}


\end{document}